%% file: root.tex
\documentclass[letterpaper, 10 pt, conference]{ieeeconf}  

\IEEEoverridecommandlockouts                              

\usepackage{amsmath} 
\usepackage{amssymb}  
\usepackage{cite}
\usepackage{amsmath,amssymb,amsfonts}
\usepackage{algorithmic}
\usepackage{graphicx}
\usepackage{textcomp}
\usepackage{subcaption}
\usepackage{xcolor}

\title{\LARGE \bf
NARF24: Estimating Articulated Object Structure for Implicit Rendering }

\author{Stanley Lewis$^{1}$, Tom Gao $^{1}$ and Odest Chadwicke Jenkins$^{1}$
\thanks{$^{1}$S. Lewis, Tom Gao, and O.C. Jenkins are with the Robotics Department,
        University of Michigan, Ann Arbor, MI 48109 \{{\tt\small stanlew, zimingg, ocj\}@umich.edu}\newline This work is supported in part by Ford Motor Company, in part by J.P. Morgan AI Research, and in part
by Amazon.}}%

\begin{document}

\maketitle
\thispagestyle{empty}
\pagestyle{empty}

\begin{abstract}

\input{sections/abstract}

\end{abstract}

\section{Introduction}

\input{sections/intro}

\section {Related Work}

\input{sections/related_work}

\section{Method}



\input{sections/method}

\section{Results}
\input{sections/results}

\section{Conclusion}

\input{sections/conclusion}



\bibliographystyle{ieeetr}
\bibliography{biblio}

\end{document}

%% file: sections/abstract.tex
Articulated objects and their representations pose a difficult problem for robots. These objects require not only representations of geometry and texture, but also of the various connections and joint parameters that make up each articulation. We propose a method that learns a common Neural Radiance Field (NeRF) representation across a small number of collected scenes. This representation is combined with a parts-based image segmentation to produce an implicit-space part localization, from which the connectivity and joint parameters of the articulated object can be estimated, thus enabling configuration-conditioned rendering.

%% file: sections/intro.tex

Articulated objects pose significant challenges for robots due to their complex degrees of freedom compared to rigid-body objects, complicating tasks like pose estimation and grasp synthesis. The scarcity of suitable datasets exacerbates these challenges. This work introduces NARF24, a parts-based approach leveraging Neural Radiance Fields (NeRFs) to estimate prismatic and revolute joint parameters for non-loopy articulated objects using minimal observed configurations. Our method processes posed RGB images alongside image-space part segmentations. We validate our approach on a real-world robot-collected dataset, and another real-world dataset with sparse segmentation supervision. We show an ablation case for a pipeline component and present results from a serial chain manipulator in a simulated environment.


%% file: sections/related_work.tex
Articulated objects remain a difficult class of objects for robots to work with. Explicit methods using a parts-based approach combining 3d mesh models and URDF (Unified Robotics Description Format) files have seen success \cite{pavlasek2020parts}. However, creating these models is laborious and difficult. Neural Radiance Fields (NeRF)\cite{mildenhall2021nerf} have also shown success in  breaking the reliance on mesh models\cite{lewis2022narf22}. Previous works such as NARF22 have continued the parts-based approach to produce NeRF representations of articulated objects. However, they still require a-priori knowledge of the articulated object's structure \cite{lewis2022narf22}. We utilize a shared-scene representation for part extraction and localization from segmentation masks. This enables traditional joint parameter estimation methods to produce a URDF model of the object's structure \cite{sturm2011probabilistic}.

Other works have utilized deeply learned approaches to infer object articulations. URDFormer \cite{chen2023urdformer} utilized simulation assets combined with known URDF models to learn an image-to-URDF transformer model. Weng el al. additionally inferred joint locations and angles from posed RGB input images via part segmentations \cite{weng2024neural}.

\begin{figure}
    \centering
    \includegraphics[width=.95\columnwidth]{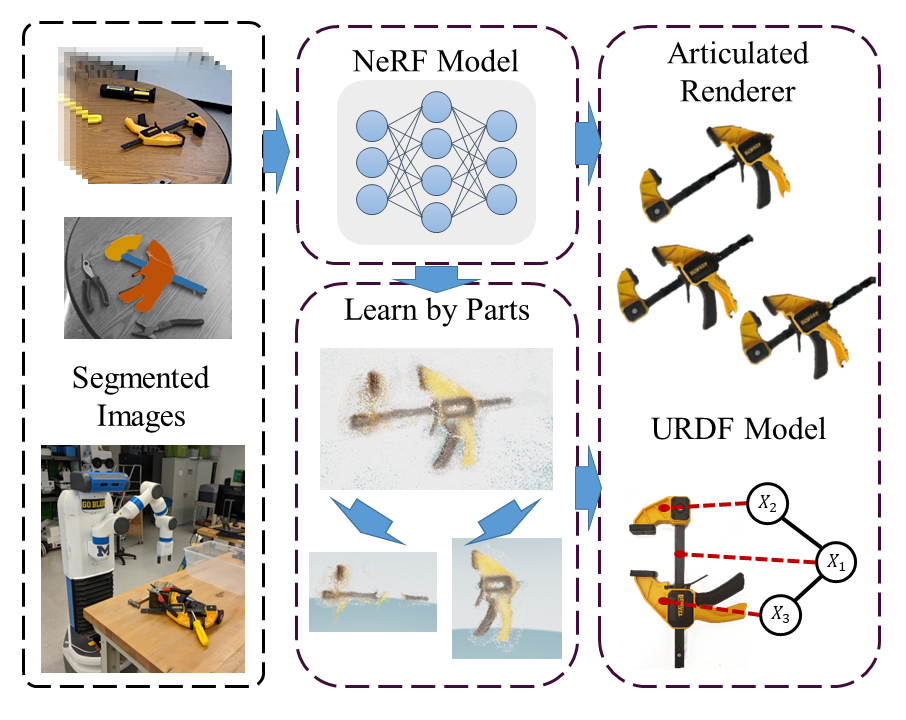}
    \label{fig:pitch_fig}
    \caption{NARF24 is a pipeline which takes in part-segmented images of an articulated object at a small number of configurations, then utilizes a scene-conditioned neural radiance field to estimate part poses and joint parameters. These create a URDF model and a subsequent articulation enabled NeRF for configurable rendering.}
\end{figure}

%% file: sections/method.tex


We start by collecting data on an articulated object at different articulation states. Each configuration example is referred to as a 'scene'. We utilize Nerfstudio \cite{tancik2023nerfstudio} to train a NeRF model that additionally contains a per-scene embedding. This embedding allows for scene-conditioned rendering and additionally makes the implicit space distribution more consistent between each scene.

We utilize the segmentation masks to create per-part point clouds within each scene. These clouds are registered to each other using Teaser++ \cite{yang2020teaser} initialized with the results of a point-to-point iterative closest point (ICP) registration.

We utilize the approach in Sturm et al. \cite{sturm2011probabilistic} to estimate the joint type and parameters between each pair of parts using the registration coordinate frames. We estimate joint connectivity and classification using a chamfer distance computed between the part point clouds, and the expected point clouds based on the predicted joint transforms. These joint predictions can then be used within the NARF22 framework \cite{lewis2022narf22} to perform configurable re-rendering of the object.

%% file: sections/results.tex
\subsection{Real World Datasets}

\subsubsection{Progress-Tools Dataset}
\begin{figure}
 \centering
 \includegraphics[width=0.9\linewidth]{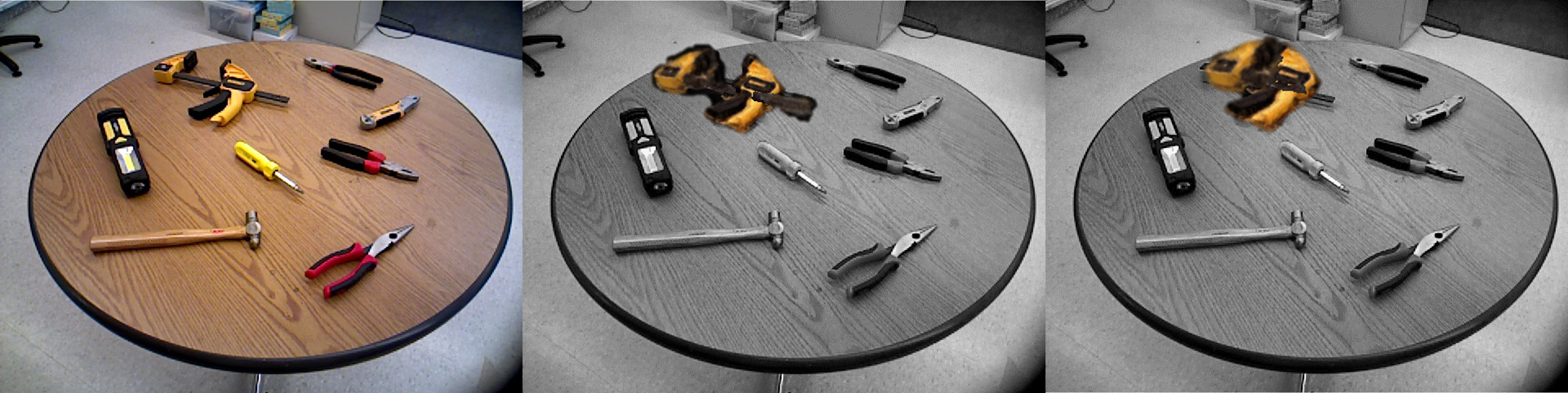}
 \caption{\label{fig:progtoolsclamp} NARF24's output when trained on the clamp in the ProgressTools dataset. \textbf{Left:} the original dataset image. \textbf{Middle:} The NARF24 output at the original pose and configuartion values. \textbf{Right:} The NARF24 output at a counterfactual configuration (fully closed).}
\end{figure}

For an initial experiment, we adopt the Progress-Tools Dataset from Pavlasek et al. \cite{pavlasek2020parts}, which contains robot-collected data on a handful of articulated tools, along with the ground truth poses and part segmentations. NARF24 is successfully able to extract the articulation estimates for the clamp, as shown in figure \ref{fig:progtoolsclamp}, which shows the learned articulated rendering at both the ground truth configuration, as well as at a counterfactual configuration as compared to the ground truth.

\subsubsection{Sparse Segments Dataset}

\begin{figure}
 \centering
 \includegraphics[width=0.9\linewidth]{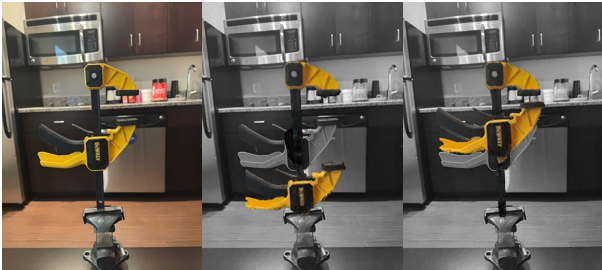}
 \caption{\label{fig:manualsparseclamp} NARF24's output when trained on the clamp with only 5 percent segmentation labeling. \textbf{Left:} The ground truth image. \textbf{Center, Right} Two counterfactual renderings of the clamp at different configurations, overlaid on the greyscale original image for context.}
\end{figure}

The most manually intensive, least scalable portion of NARF24's inputs to obtain are the part segmentations. To test how well our approach performs when only a small number of these segmentations are provided, a dataset for the clamp was collected, but only 5\% of the images were labelled with segmentation masks. Images without masks were used to train the shared-scene representation, but were not utilized in the parts separation or subsequent registration steps. Even with the lower segmentation coverage, figure \ref{fig:manualsparseclamp} shows that acceptable articulation estimates were obtained.

\subsubsection{Registration Ablation}

\begin{figure}
    \centering
    \includegraphics[width=.9\linewidth]{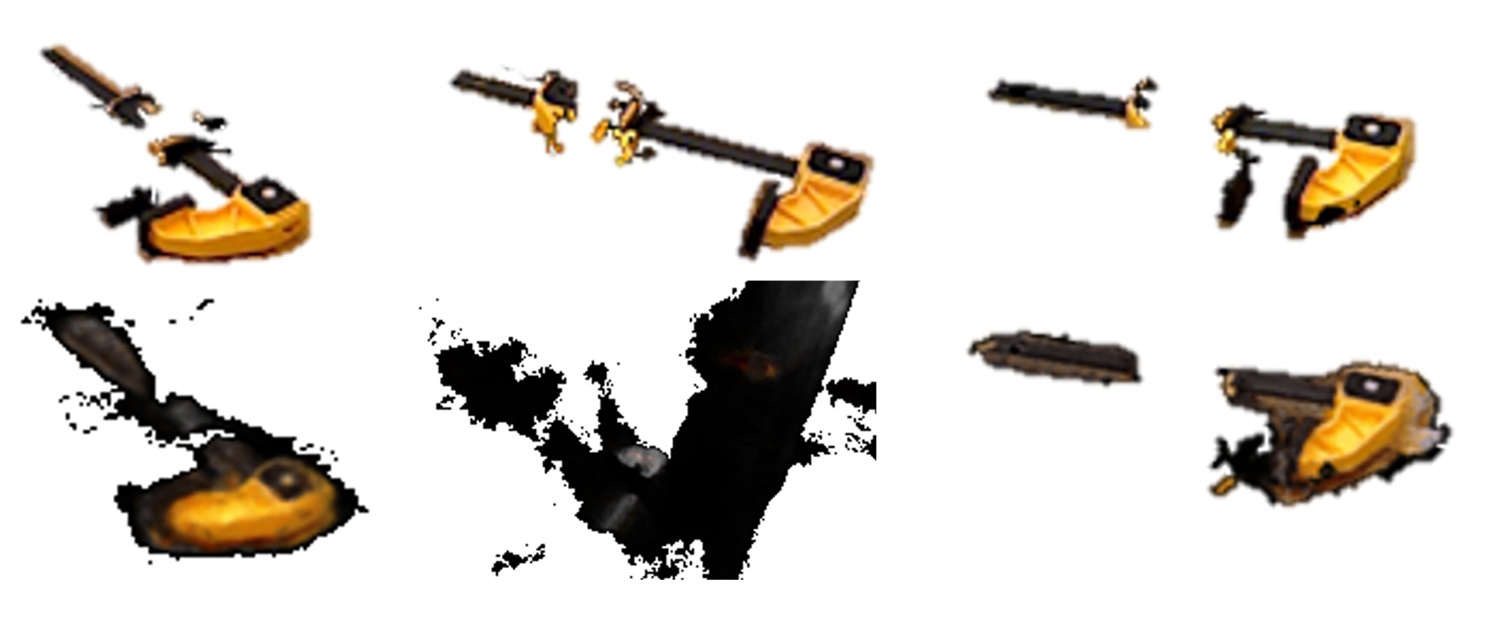}
    
 \caption{\label{fig:ablation}Qualitative ablation study on training a single-part NeRF subsequent to the part registration step. Top is the ground truth part, and bottom is the NeRF rendering. \\\hspace{\textwidth}
 \textbf{Left (ICP only):} Performs adequately.\\\hspace{\textwidth}
 \textbf{Center (Teaser++ only):} fails for single-part NeRF training. \\\hspace{\textwidth}
 \textbf{Right (ICP \& Teaser++):} Produces the best output.}
\end{figure}

The part registration step is the most sensitive part of the NARF24 pipeline with respect to estimating object structure. Poor per-part localization leads to inaccurate joint estimations, which hinders any downstream parts-based training process. We thus performed an ablation study to show that our ICP initialized Teaser++ method is effective. Figure \ref{fig:ablation} shows the resulting point clouds after training a single-part NeRF subsequent to estimating cross-scene part registration with ICP only, Teaser++ only, and Teaser++ with ICP initialization.

\subsection{Simulated Articulated Arm}
\begin{figure}
    \centering
    \begin{subfigure}{0.45\linewidth}
        \includegraphics[width=\textwidth]{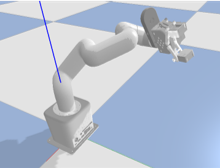}
    \end{subfigure}
    \begin{subfigure}{0.45\linewidth}
        \includegraphics[width=\textwidth]{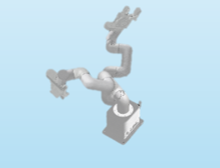}
    \end{subfigure}
 \caption{\label{fig:mycobot}\textbf{Left:} Example output from the simulator environment. \textbf{Right:} Renderings of the arm at two different configurations of every joint, after training on the generated sim data (overlaid at base part)}
\end{figure}

In order to demonstrate the best-case capabilities of the NARF24 system, a simulated dataset was created in PyBullet of a MyCobot 6 Degree of Freedom robot arm. Due to its simulated nature, this dataset has perfect camera poses, part segmentations, and part poses. Figure \ref{fig:mycobot} shows an example rendering from the sim environment, along with a pair of renderings overlaid on top of each other, showing each of the arm's joints being changed.

%% file: sections/conclusion.tex
NARF24 adopts a parts-based approach to enable more scalable learning and rendering of NeRF based implicit models for articulated objects. Results show that real-world data can be used to generate configuration-conditioned renderers even with small amounts of segmentation labels, and a simulated data example shows the effectiveness on a complex robot arm.

%% file: root.bbl
\begin{thebibliography}{1}

\bibitem{pavlasek2020parts}
J.~Pavlasek, S.~Lewis, K.~Desingh, and O.~C. Jenkins, ``Parts-based articulated object localization in clutter using belief propagation,'' in {\em 2020 IEEE/RSJ International Conference on Intelligent Robots and Systems (IROS)}, pp.~10595--10602, IEEE, 2020.

\bibitem{mildenhall2021nerf}
B.~Mildenhall, P.~P. Srinivasan, M.~Tancik, J.~T. Barron, R.~Ramamoorthi, and R.~Ng, ``Nerf: Representing scenes as neural radiance fields for view synthesis,'' {\em Communications of the ACM}, vol.~65, no.~1, pp.~99--106, 2021.

\bibitem{lewis2022narf22}
S.~Lewis, J.~Pavlasek, and O.~C. Jenkins, ``Narf22: Neural articulated radiance fields for configuration-aware rendering,'' in {\em 2022 IEEE/RSJ International Conference on Intelligent Robots and Systems (IROS)}, pp.~770--777, IEEE, 2022.

\bibitem{sturm2011probabilistic}
J.~Sturm, C.~Stachniss, and W.~Burgard, ``A probabilistic framework for learning kinematic models of articulated objects,'' {\em Journal of Artificial Intelligence Research}, vol.~41, pp.~477--526, 2011.

\bibitem{chen2023urdformer}
Q.~Chen, M.~Memmel, A.~Fang, A.~Walsman, D.~Fox, and A.~Gupta, ``Urdformer: Constructing interactive realistic scenes from real images via simulation and generative modeling,'' in {\em Towards Generalist Robots: Learning Paradigms for Scalable Skill Acquisition@ CoRL2023}, 2023.

\bibitem{weng2024neural}
Y.~Weng, B.~Wen, J.~Tremblay, V.~Blukis, D.~Fox, L.~Guibas, and S.~Birchfield, ``Neural implicit representation for building digital twins of unknown articulated objects,'' in {\em Proceedings of the IEEE/CVF Conference on Computer Vision and Pattern Recognition}, pp.~3141--3150, 2024.

\bibitem{tancik2023nerfstudio}
M.~Tancik, E.~Weber, E.~Ng, R.~Li, B.~Yi, J.~Kerr, T.~Wang, A.~Kristoffersen, J.~Austin, K.~Salahi, {\em et~al.}, ``Nerfstudio: A modular framework for neural radiance field development,'' {\em arXiv preprint arXiv:2302.04264}, 2023.

\bibitem{yang2020teaser}
H.~Yang, J.~Shi, and L.~Carlone, ``Teaser: Fast and certifiable point cloud registration,'' {\em IEEE Transactions on Robotics}, vol.~37, no.~2, pp.~314--333, 2020.

\end{thebibliography}
